\documentclass[letterpaper, table]{article}  
\usepackage{aaai2026}  
\usepackage{times}  
\usepackage{helvet}  
\usepackage{courier}  
\usepackage[hyphens]{url}  
\usepackage{graphicx} 
\usepackage{amsfonts}
\usepackage{amsmath}
\usepackage{pifont}
\usepackage{xcolor} 
\usepackage{natbib}  
\usepackage{caption} 
\usepackage{subcaption}
\usepackage{float} 
\usepackage{makecell}
\usepackage{algorithm}
\usepackage{algorithmicx}  
\usepackage{algpseudocode} 
\usepackage{booktabs}
\usepackage{multirow}
\usepackage{threeparttable}
\floatname{algorithm}{Algorithm}  

\usepackage{newfloat}
\usepackage{listings}
\DeclareCaptionStyle{ruled}{labelfont=normalfont,labelsep=colon,strut=off} 
\floatstyle{ruled}
\newfloat{listing}{tb}{lst}{}
\floatname{listing}{Listing}
\title{SAGE: A Lightweight Framework for Trigger-Guided LoRA-Based Self-Adaptation in LLMs}

\author{
    Jiacheng Wei, Faguo Wu, Xiao Zhang
}
\affiliations{
    School of Artificial Intelligence, Beihang University, Beijing\\
    Beijing Advanced Innovation Center for Future Blockchain and Privacy Computing, Beihang University, Beijing\\
    \texttt{jakiewei258@gmail.com}
}

\usepackage{bibentry}

\begin{document}

\maketitle

\begin{abstract}
Large language models (LLMs) are unable to continuously adapt and learn from new data during reasoning at inference time. To address this limitation, we propose that complex reasoning tasks be decomposed into atomic subtasks and introduce SAGE, a trigger-guided dynamic fine-tuning framework that enables adaptive updates during reasoning at inference time. SAGE consists of three key components: (1) a Trigger module that detects reasoning failures through multiple evaluation metrics in real time; (2) a Trigger Buffer module that clusters anomaly samples using a streaming clustering process with HDBSCAN, followed by stability checks and similarity-based merging; and (3) a Lora Store module that dynamically optimizes parameter updates with an adapter pool for knowledge retention. Evaluation results show that SAGE demonstrates excellent accuracy, robustness, and stability on the atomic reasoning subtask through dynamic knowledge updating during test time. Specifically, an EM accuracy of $97.16\%_{\pm 4.65\%}$ reflects statistically significant and reliable performance.
\end{abstract}

\section{Introduction}
\begin{figure}[t]
\centering
\includegraphics[scale=0.3]{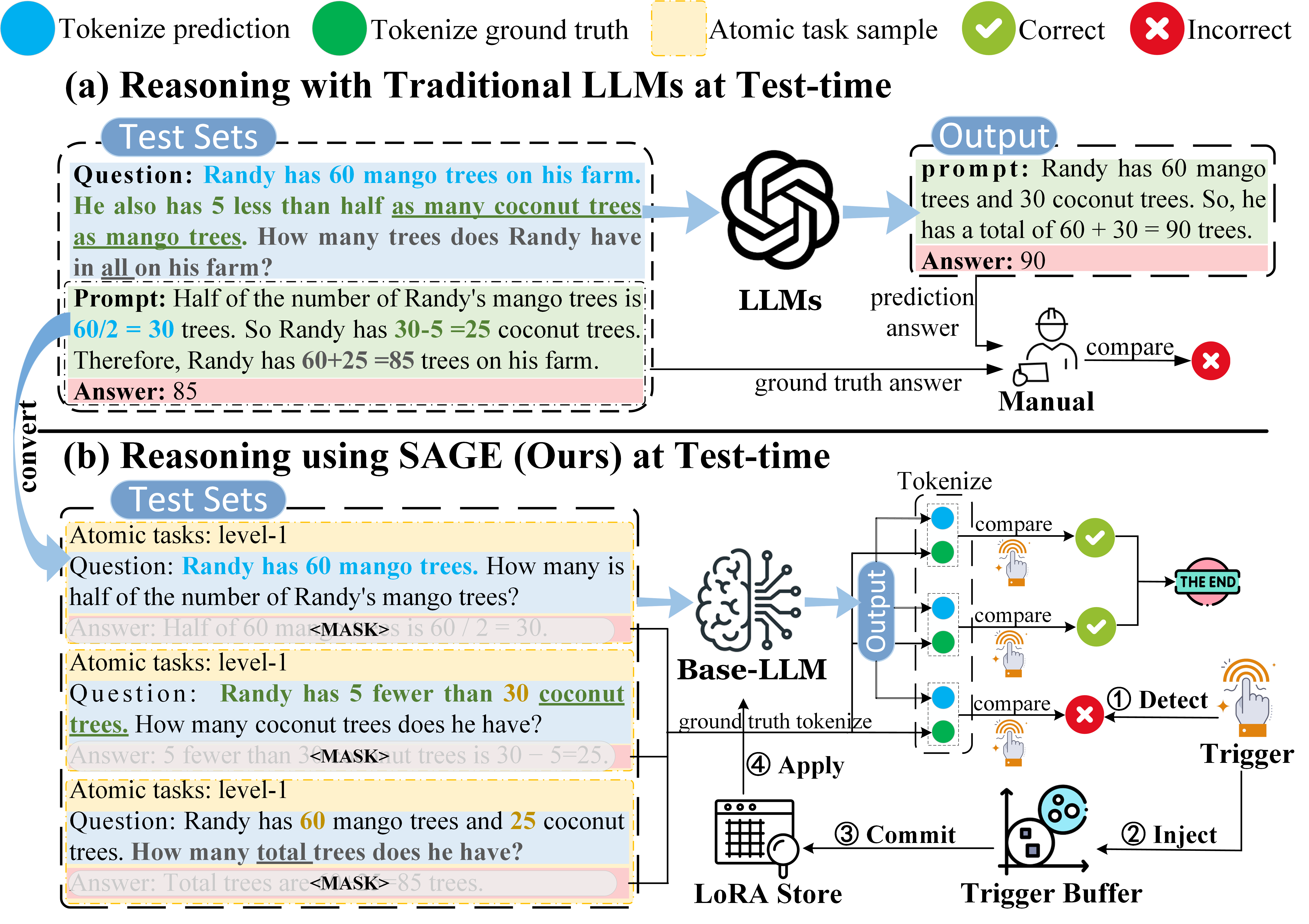}
\caption{Comparison of reasoning in traditional LLMs and SAGE at test-time. (a) Traditional LLMs struggle with complex reasoning, where tasks are entangled, and errors require manual verification. (b) SAGE decomposes reasoning into atomic tasks (yellow numbers indicate answers from previous steps). The Trigger module detects anomalies by comparing predictions with ground truth, injects data into the Trigger Buffer for clustering, and commits the clusters to the LoRA Store for fine-tuning. The resulting adapter is finally applied to improve LLM performance.}
\label{complete}
\end{figure}

Large language models (LLMs) have demonstrated remarkable generalization capabilities by pre-training on massive text corpora, but they are unable to gradually learn and adapt from new data during reasoning (as shown in Figure~\ref{complete}(a)). This limitation~\cite{dziri2023faith, kil2024mllm, jin2025reasoning} prevents LLMs from handling tasks that require adaptation to new environments or changes over time.  As we move toward true general artificial intelligence (AGI), enabling LLMs to support real-time adaptation and continual learning becomes increasingly important. In particular, reasoning tasks inherently require the integration of new knowledge and adaptation to evolving contexts, making dynamic updates crucial. Therefore, this paper aims to explore the following core question: \textit{How can LLMs adaptively and incrementally update themselves to cope with reasoning failures?}

The adaptive update task of LLMs is essentially similar to the traditional post-training update, but it requires dynamic automatic updates during reasoning in the inference phase. To understand this task, we need to review the existing update paradigms. Currently, the mainstream update methods for LLMs are divided into two categories: static fine-tuning and external enhancement. Static fine-tuning usually adjusts the model's behavior by combining Supervised Fine-Tuning (SFT) and Reinforcement Learning (RL)~\cite{ouyang2022training, tie2025survey}. However, if reasoning task can be decomposed into atomic subtasks, reinforcement learning is no longer the best solution. Lightweight methods such as low-rank adaptation (LoRA)~\cite{hu2022lora} are more suitable, as they can perform targeted updates without retraining the entire model. Although external enhancement methods (such as retrieval-augmented generation (RAG)~\cite{lewis2020retrieval} and long-context prompting) can alleviate this problem, RAG usually lack deep integration and persistent memory and long-context prompting have limited performance in handling reasoning failures caused by new knowledge transfer. In light of these limitations, we argue that \textbf{adaptive updates should occur during test-time training, with task objectives focused on integrating new knowledge similar to post-training, while self-adaptation are optimized via reasoning subtasks to meet dynamic requirements}.

To this end, we propose \textbf{SAGE} in Figure~\ref{complete} (b), a trigger-guided dynamic fine-tuning framework designed to achieve adaptive updates of LLMs on reasoning subtasks at test time. SAGE implements three core functions: (1) detecting reasoning failures, (2) clustering anomaly samples, and (3) dynamically optimizing parameter updates. Together, these modules form a lightweight mechanism that enables LLMs to localize and adapt to failures caused by underrepresented or emerging knowledge, without full model retraining. The Trigger module evaluates model outputs across multiple dimensions (surface text, model behavior, and semantic representation levels) to detect reasoning failures in real time without additional supervision. The Trigger Buffer module then clusters these anomaly samples, first bucketing them according to the task to reduce domain interference, and then applying a streaming clustering process: initial HDBSCAN clustering and cluster stability check (e.g. Adjust Rank Index, cosine embedding), followed by merging through embedding similarity. Finally, the Lora Store module maintains a dynamic pool of parameter-efficient adapters. It initially searches through parameter space (e.g., rank, learning rate, dropout) for adapter training, ranks Top-$k$ candidate configurations by accuracy and loss, and then conducts local expansion optimization to retain the Top-$3$ adapters for reuse.

In the experiments, each module was evaluated as follows: The Trigger module was assessed through false positive rate detection, threshold sensitivity analysis, and indicator weight sensitivity analysis. The results demonstrate the module's strong discriminative power in out-of-distribution datasets, validating the effectiveness of the selected indicators. Specifically, approximately 57\% of the plateaus occurred within the threshold range, while 62.5\% occurred within the indicator weight range, indicating robust accuracy under varying conditions. For the Trigger Buffer module, ablation studies on HDBSCAN, stability checks, and merging confirmed the design's validity. Dynamic visualization and sensitivity analysis of stream data clustering demonstrate reliable performance with streaming data. The evaluation of the LoRA Store module focused on fine-tuning accuracy. In atomic task evaluations, it outperformed reasoning with multitask datasets by more than 50\%. Additionally, heatmap tests of LoRA parameter rank and learning rate show significant impact on fine-tuning accuracy (up to 83\%). After combining the three modules, the EM accuracy of the SAGE framework increased from 81.91\% to 94.85\%, while the MAE and MSE decreased by several orders of magnitude, from an average of $10^7$ to 0.1, demonstrating enhanced robustness and stability. Finally, the ablation study further validated SAGE's design and its effectiveness in supporting reasoning during inference.

In summary, the contributions of SAGE framework are shown as follows:
\begin{itemize}
    \item We define the challenge of real-time self-adaptation for LLMs in streaming data environments, and propose replacing RL-based post-training adaptation with atomic reasoning subtasks, which decouple hyperparameter dependencies, enabling a LoRA-only, statically compatible continual update mechanism. 
    \item We introduce a trigger-based adaptation methods, where reasoning failures serve as natural signals for model update. This transforms static LoRA updates into event-driven learning, enabling LLMs to self-adaptation based on inference-time feedback.
    \item We present SAGE, a lightweight architecture composed of a Trigger module for anomaly detection, a Trigger Buffer module for clustering anomaly data, and a LoRA Store module for efficient fine-tuning. The Top-$3$ LoRA adapters are selected by performance and reused for future reasoning tasks.
    \item Extensive experiments show that SAGE achieves high accuracy and robustness across all modules. With atomic subtask partitioning, fine-tuned LoRA adapters consistently achieve more than 80\% accuracy, with some tasks exceeding 99\%.
\end{itemize}

\section{Preliminaries}
\paragraph{Problem Statement}
The core challenge addressed in this work is \textit{designing self-adaptive LLMs capable of continuously incorporating new inputs at test-time, ensuring robust knowledge integration despite limited supervision and sparse data}. To clarify this challenge, we formulate the following design requirements for self-adaptive LLMs: \textbf{Adaptation scope}: The model should \textit{perform localized parameter updates}, avoiding large-scale parameter updates. \textbf{Adaptation phase}: Training must occur \textit{during test-time interaction}, as the model engages with users in real-time to incorporate new knowledge, rather than during post-training or pre-training of LLMs. \textbf{Adaptation objective}: The model must ensure that \textit{updated parameters preserve accuracy and robustness}, preventing undesirable drift in global parameters. \textbf{Adaptation autonomy}: The model should \textit{autonomously determine when and how to adapt}, selecting relevant instances from interactive data streams,  while independently optimizing algorithms and hyperparameters without external supervision. \textbf{Adaptation challenges}: The model mustfunction in environments where data is \textit{noisy, limited, and unordered}, with incoming tasks that are open-ended and dynamically evolving.

\paragraph{Sketch of SAGE}
\begin{figure*}[t]
\centering
\includegraphics[scale=0.32]{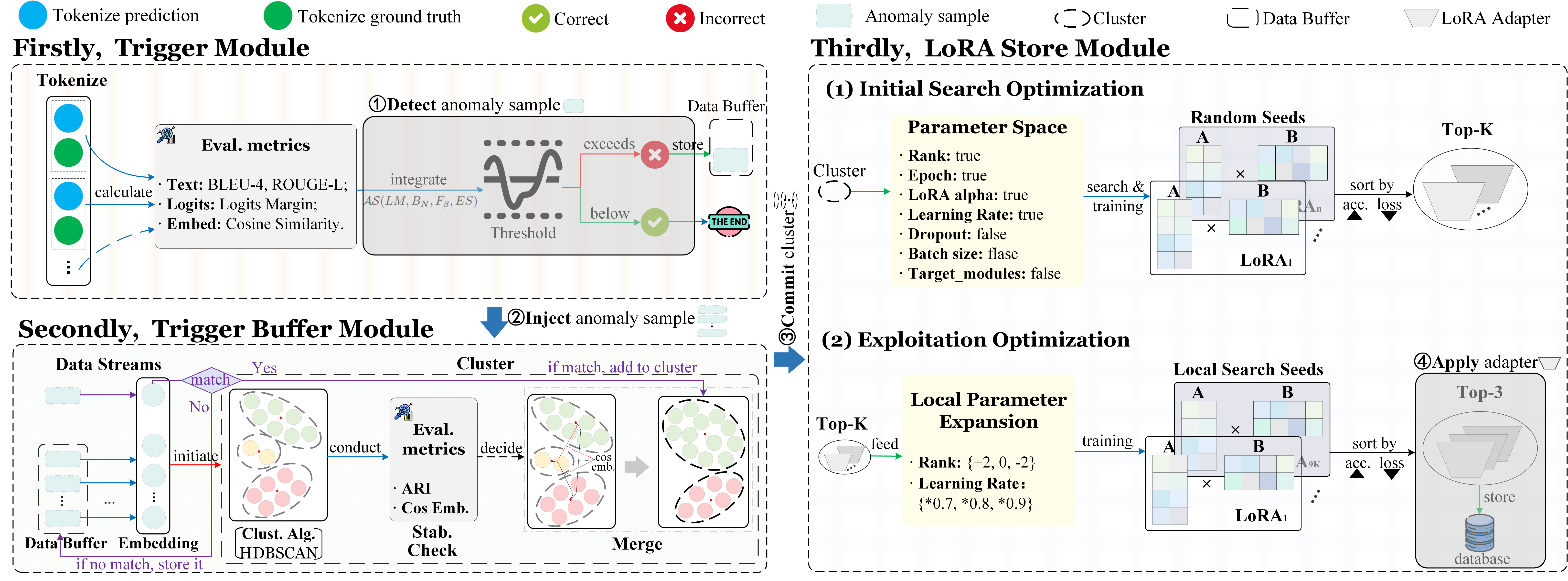}
\caption{Overview of the SAGE architecture. (a) The Trigger module evaluates predictions across surface text,  model behavioral, and semantic  representation levels, normalizes the scores, and performs threshold-based anomaly detection. (b) The Trigger Buffer collects streaming anomalies, clusters them via HDBSCAN, followed by stability check and similarity-based merging. (c) The LoRA Store samples parameter space for adapter training, ranks Top-k adapters by accuracy and cross-entropy loss, then conducts local expansion over rank and learning rate, and retains the Top-3 adapters for fine-tuning.}
\label{framework}
\end{figure*}

To address the challenges of self-adaptive LLMs, we propose a trigger-guided self-adaptation framework based on lightweight LoRA tuning, a parameter-efficient technique that avoids full model retraining. Our key insight is to reduce adaptation complexity through \textit{decomposing complex reasoning tasks into atomic subtasks}, which are minimal and independent, enabling the model to adapt more easily, reducing error accumulation and ensuring stable updates for improved accuracy. Based on this insight, we introduce \textbf{SAGE}, a modular framework consisting of the following components: \textbf{Trigger}, which converts static fine-tuning into dynamic adaptation by detecting reasoning failures on out-of-distribution inputs and determines whether the data should be retained for future adaptation; \textbf{Trigger Buffer}, which caches and cluster anomalous samples to improving the quality of subsequent fine-tuning by increasing inter-sample coherence; and \textbf{LoRA Store}, which performs parameter-efficient fine-tuning on stable clusters, with parameters automatically adjusted based on performance. The most effective adapters are preserved for future reuse, ensuring efficient adaptation in similar tasks.

\section{Design of SAGE}
\subsection{Trigger}
The Trigger module is responsible for initiating dynamic adaptation by detecting reasoning failures of LLMs. As shown in Figure~\ref{complete}(b), given a test set, the LLM first receives a masked version of the gold answer, which is retained as the ground truth. It then generates an answer based on the question, producing the predicted answer. The Trigger module compares the ground truth with the predicted answer at the token level to determine whether a reasoning failure has occurred. To support this comparison, we use three evaluation level in Figure~\ref{framework}: surface text (\textit{Logits Margin}),  model behavioral (\textit{BLEU} and \textit{ROUGE-L}), and semantic representation levels (\textit{Embedding Similarity}).

\textit{Logits Margin}. Margin~\cite{liu2016large} is used to evaluate the confidence of LLMs in next-token prediction. Specifically, the logits margin $LM$ indicates whether a prediction is in-distribution. Assume the model's output logits form a vector $z \in \mathbb{R} ^V$, where $V$ is the vocabulary size. Then the logits margin can be expressed as,
\begin{equation}
\label{margin}
    LM= \max_{i} z_i-\max_{j\ne i} z_j
\end{equation}

\textit{BLEU}. Bilingual Evaluation~\cite{papineni2002bleu} $B_n$ is a metric originally designed for machine translation quality. We adopt the BLEU-$n$ variant with a brevity penalty \text{BP} to measure the n-gram overlap between predicted and ground truth answers, defined in terms of n-gram precision $p_n$ and weights $w_n$ for each n-gram as follows,

\begin{equation}
\label{BLEU}
    B_{n} = \text{BP} \cdot \exp\left( \sum_{n=1}^{N} w_n \log p_n \right)
\end{equation}

\textit{ROUGE-L}. ROUGE-L~\cite{lin2004rouge} and BLEU both evaluate the similarity between a predicted answer and the ground truth. While BLEU focuses on contiguous n-gram matches to capture local accuracy, ROUGE-L emphasizes structural similarity by measuring the longest common subsequence regardless of contiguity. We adopt the $F_1$ score, computed as the harmonic mean of precision $P$ and recall $R$, and define it as follows,

\begin{equation}
\label{rouge-L}
    F_{\beta} = \frac{(1 + \beta^2) \cdot P \cdot R}{\beta^2 \cdot P + R}
\end{equation}

\textit{Embedding Similarity}. Embedding Similarity~\cite{lin2017structured} $ES$ measures the semantic similarity between the predicted answer and the ground truth answer. Let $u_{pred}$, $v_{true}$ denote the embedding vectors of the predicted and ground truth answers, respectively. The formula is,

\begin{equation}
\label{emb_similarity}
    \text{ES} = \frac{\mathbf{u_{pred}} \cdot \mathbf{v_{true}}}{\|\mathbf{u_{pred}}\| \cdot \|\mathbf{v_{true}}\|}
\end{equation}

To obtain the final Anomaly Score $\mathcal{AS}$, where higher values indicate greater likelihood, we perform normalization and integration of Equation~\eqref{margin}, Equation~\eqref{BLEU}, Equation~\eqref{rouge-L} and Equation~\eqref{emb_similarity} as follows,

\begin{equation}
    \mathcal{AS}(LM, B_N, F_\beta, ES) = \sum_{i=1}^{4} w_i(1 - s_i)
\end{equation}

where $s_1 = LM/LM_m$, $s_2 = B_4$, $s_3 = F_1$, and $s_4 = ES$. Here, $LM_m$ denotes the upper bound used to normalize the margin score, which is set to 5.0 in our experiments. The weight $w_i$ corresponds to each evaluation metric. Finally, a threshold is applied to the computed anomaly score, and any sample exceeding it is flagged as an anomaly. To further reduce false positives, the score is attenuated when the embedding similarity falls within a reference percentile range derived from in-distribution data.

\subsection{Trigger Buffer}
The Trigger Buffer module clusters anomalous samples. Despite established clustering techniques, two challenges persist: \ding{192} \textbf{limited data}, due to SAGE's dynamic triggering at test time with small, atomic tasks; and \ding{193} \textbf{streaming data}, which arrives incrementally rather than in bulk, complicating clustering. To address the above challenges, we designed the Streaming Buffer Clustering (SBC) algorithm in Appendix A.1 and the overall process is shown in Figure~\ref{framework}.

For each newly arrived abnormal sample $x$, the SBC algorithm infers its structure tag and extracts its semantic embedding vector and keyword set for subsequent similarity matching (violet line Figure~\ref{framework}). If the cluster for the  structure tag $s$ is deemed "\textit{stable}" (i.e., a structure cluster $\mathcal{C}_{s}$ already exists), SBC algorithm computes the similarity between the current sample and all candidate clusters under the same tag, selecting the one with the highest score (Line~5). The clustering score $\gamma$ is a weighted combination of semantic embedding similarity and keyword overlap. If $\gamma$ exceeds the threshold $\tau$, the sample $x$ is added to the selected cluster $\mathcal{C}^{\ast}$ (Line~8), and the assignment result is returned, terminating the process. If the structure cluster is not stable or the score does not meet the threshold, the sample is temporarily stored in the buffer $\mathcal{B}_{s}$ for tag $s$ (Line~12), awaiting further processing by the delayed clustering mechanism.

When a newly abnormal sample $x$ causes the number of samples in the buffer associated with structure tag $s$ to reach the predefined threshold $T$ (Line~14), the SBC algorithm triggers batch clustering. The system extracts the semantic embedding vectors ${\mathbf{e}_i}$ of all buffered samples (Line~15), then applies \textbf{HDBSCAN}~\citep{campello2013density} for unsupervised density-based clustering. HDBSCAN is preferred over algorithms like $k$-means, which require preset cluster numbers, for two reasons: \ding{192} it adapts to dynamic, unknown cluster structures in streaming data, and \ding{193} its density-based approach is robust to small, non-uniform datasets, identifying high-density regions effectively.

However, HDBSCAN alone cannot guarantee clustering \emph{stability}. To address this, we introduce two stability check metrics: \ding{192} the \textbf{Adjusted Rand Index (ARI)}, measuring label consistency across successive clusterings; and \ding{193} the \textbf{Average Cosine Similarity}, quantifying semantic coherence between new and previous cluster centroids. When both exceed their thresholds $\eta_{\text{ARI}}$ and $\eta_{\cos}$, clustering is considered stable. The buffered samples are migrated to a newly formed formal cluster $\mathcal{C}_{\text{new}}$, the buffer $\mathcal{B}_s$ cleared (Line~21), and the structure tag $s$ marked as "\textit{stable}" (Line~22). 

Given that streaming input can fragment clusters, when the number of stable clusters for tag $s$ exceeds the threshold, an \textbf{inter-cluster merge} (Line~24) is triggered. Clusters are merged if their centroids’ cosine similarity exceeds threshold $\delta$, or ARI shows high label consistency, reducing redundancy and increasing sample density. If batch clustering fails to meet stability criteria, the result is discarded and samples remain buffered for future clustering(Line~28). Finally, samples unassigned to any cluster and without triggered clustering are marked as "\textit{unassigned}" until clustering conditions are met (Line~31).

\subsection{Lora Store}
The LoRA Store module is responsible for fine-tuning the stable clusters in the Trigger Buffer. There are also two major challenges: \ding{192} different types of datasets require specific tuning to obtain the optimal configuration; \ding{193} streaming inputs may alter the optimal adapter over time, requiring the storage of multiple adapters to accommodate dynamic changes. To efficiently fine-tune parameters under limited data and diverse atomic tasks, we propose Cluster-Aware LoRA Optimization (CLO), a clustering-based method that automatically searches for the optimal LoRA configuration $\theta^*$, saving the corresponding adapter for subsequent inference or combination (see Appendix A.2) and the overall process is shown in Figure~\ref{framework}

\paragraph{Initial Search optimization}
Given a cluster $\mathcal{C}$ with a stable structure, CLO algorithm randomly samples $n$ candidate configurations from the hyperparameter search space $\mathcal{P}$ to form the initial configuration set $\Theta_0$.  Simultaneously, the system creates an empty list to store the results of each configuration after training, including validation accuracy, cross-entropy loss, save path, and corresponding LoRA adapter configuration information. Then, CLO algorithm performs complete LoRA fine-tuning on each configuration in $\Theta_0$ for the samples in the cluster $\mathcal{C}$ (Line~4), recording the training results to form a preliminary result set $\mathcal{R}_0$ (Line~5). Next, CLO algorithm ranks the results in $\mathcal{R}_0$ based on a sorting strategy that prioritizes validation accuracy firstly and cross-entropy loss secondly, selecting the Top-$k$ configurations (Line~9). This strategy accounts for the limited data size, prioritizing accuracy to ensure SAGE's correct response to similar inputs. 

\paragraph{Exploitation optimization}
Based on these $k$ optimal configurations, CLO algorithm conducts exploitation optimization in their neighboring parameter spaces (Line~12), mainly adjusting the LoRA rank and learning rate—parameters with the greatest impact on fine-tuning across diverse reasoning domains. For each configuration in the fine-tuned set $\Theta_{\text{opt}}$, CLO algorithm continues LoRA fine-tuning,  resulting in a refined result set $\mathcal{R}_1$ (Line~14). Finally, CLO algorithm merges the preliminary result set $\mathcal{R}_0$ with the refined set $\mathcal{R}_1$ into the total result set $\mathcal{R}{\text{all}}$ (Line~19), sorts it again by accuracy and cross-entropy loss, and selects the final Top-3 configurations (Line~21). The corresponding LoRA adapters are saved as the final optimized results $\mathcal{A}_{\text{LoRA}}$ (Line~22), and the three optimal adapters, along with the best configuration $\theta^*$, are returned for subsequent reasoning calls (Line~23). For new data with different characteristics, adaptation is achieved by triggering additional LoRA fine-tuning. 

The CLO algorithm combining initial exploration with local refinement, making it suitable for atomic tasks characterized by sparse data and diverse atomic tasks. It ensures that each cluster has the optimal LoRA configuration for the current data, providing a solid foundation for subsequent adapter upgrades and dynamic fine-tuning.

\section{Experiments}
\subsection{Experimental Settings}
All experiments are conducted on a single NVIDIA A100 40GB GPU using LLaMA-2-7B as the base model. Parameter-efficient fine-tuning is performed via LoRA across all settings. For trigger detection and clustering, we employ BGE-large-en-v1.5 as the embedding model. Evaluation is conducted on four datasets designed to capture both in-distribution (ID) and out-of-distribution (OOD) behaviors. We use TriviaQA as the ID benchmark, which matches the model’s pretraining distribution. To evaluate generalization under knowledge gaps, we include three OOD datasets from distinct domains: PubMedQA (biomedical Q\&A), LexGlue (legal summarization), and GSM8K (mathematical reasoning). These datasets are chosen to represent structurally diverse, low-coverage regions unlikely to be encountered during pretraining.

\subsection{Module-wise Evaluation of SAGE}
\paragraph{Evaluation of Trigger Detection Accuracy.}

\begin{figure*}[t]
    \centering
    \begin{subfigure}[b]{0.24\textwidth}
        \includegraphics[width=\linewidth]{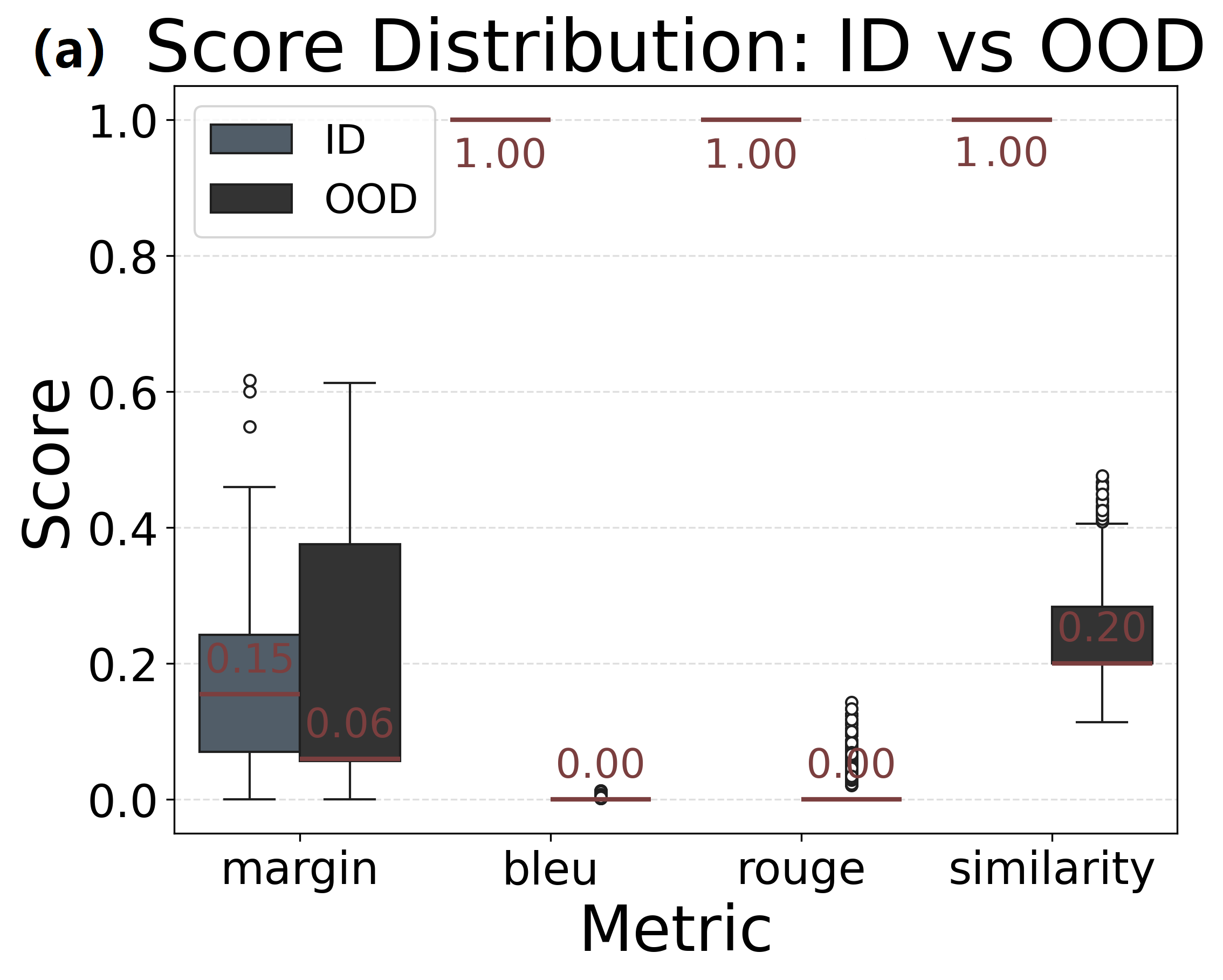}
    \end{subfigure}
    \hfill
    \begin{subfigure}[b]{0.24\textwidth}
        \includegraphics[width=\linewidth]{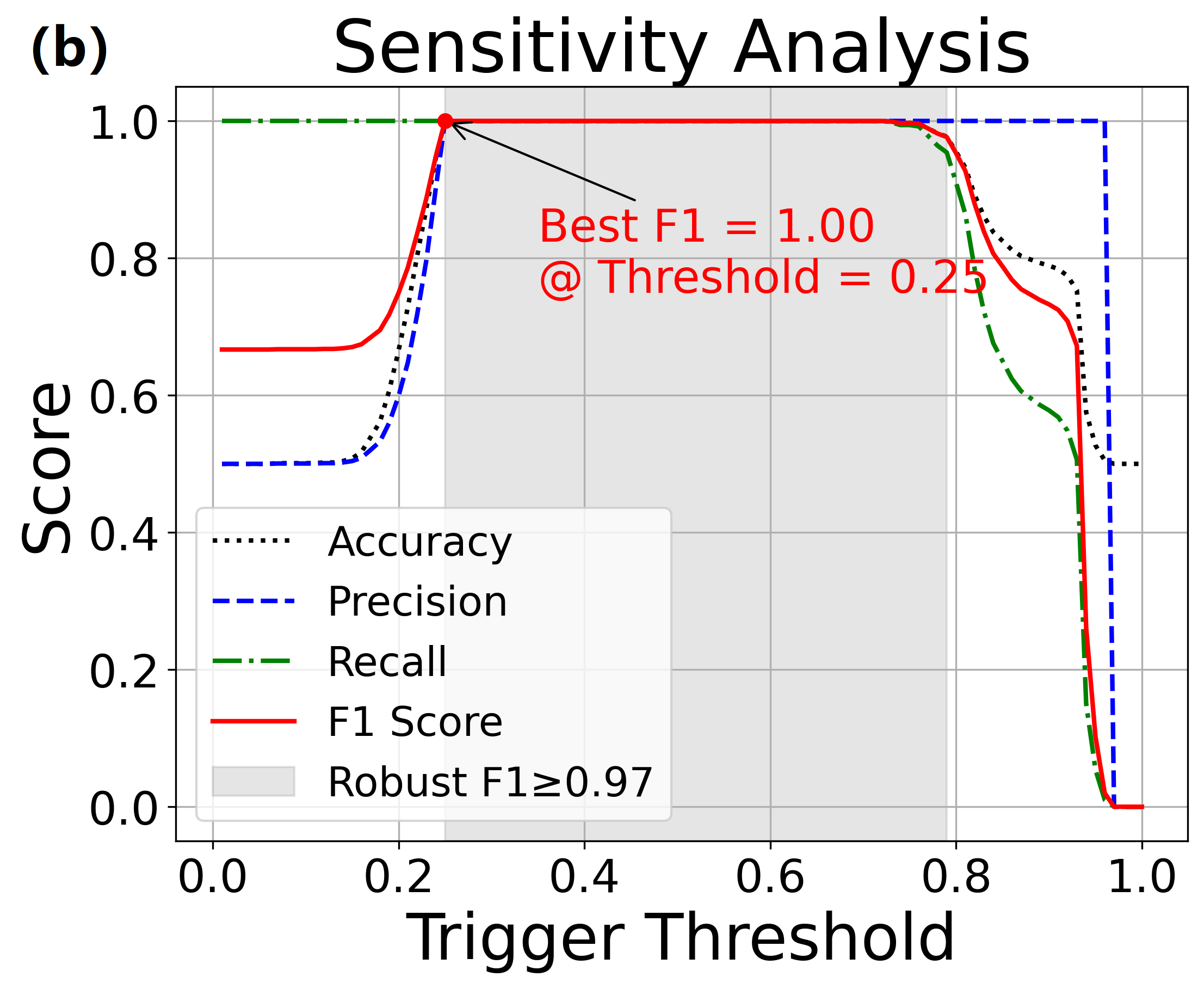}
    \end{subfigure}
    \hfill
    \begin{subfigure}[b]{0.48\textwidth}
        \includegraphics[width=\linewidth]{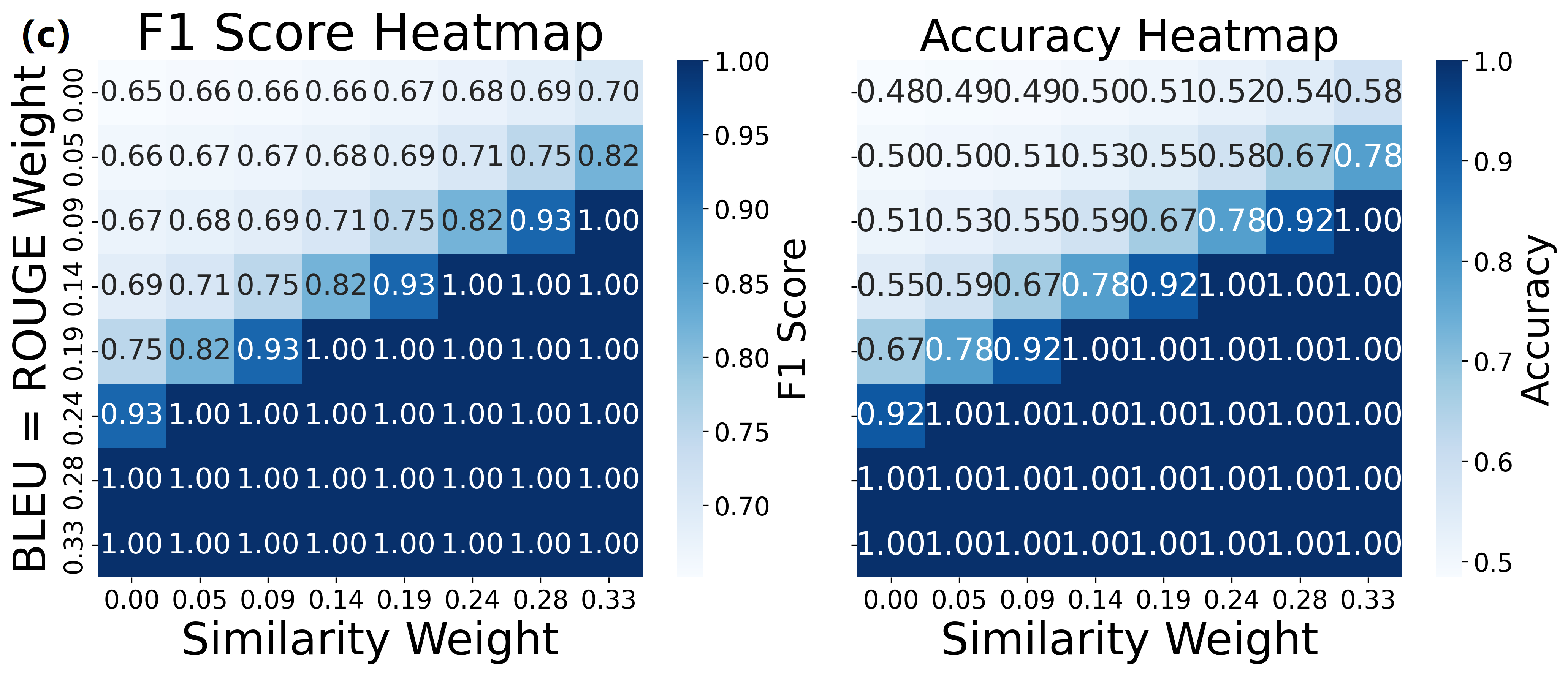}
    \end{subfigure}

    \caption{Parameter analysis of the Trigger Module. (a) False positive rates on ID vs. OOD samples. Red lines in boxplots indicate the mean; the logits margin metric is normalized. (b) Threshold sensitivity analysis; the gray area marks the optimal threshold range. (c) Heatmap of score weight sensitivity. The logits margin weight is computed dynamically from BLEU, ROUGE-L, and embedding similarity differences.}
    \label{fig:tri-results}
\end{figure*}

To fairly evaluate the detection capability of the Trigger module, we constructed a mixed test set containing equal proportions of ID and OOD samples. The OOD samples were drawn in equal parts from three domain-specific datasets: PubMedQA, LexGlue, and GSM8K. All samples were uniformly converted into a structure consisting of the fields "\textit{question}", "\textit{full prompt}", "\textit{real-answer}", and "\textit{label}", where \textit{label = 0} denotes ID samples and \textit{label = 1} denotes OOD samples.

To evaluate the Trigger module’s accuracy and robustness, we conducted three experiments (Figure~\ref{fig:tri-results}). Figure~\ref{fig:tri-results}(a) shows that semantic indicators (BLEU, ROUGE, similarity) score near 1 on ID samples and significantly lower on OOD, indicating strong discriminative power. Logits margin shows smaller variance but clear mean differences. Figure~\ref{fig:tri-results}(b) demonstrates stable performance across a broad threshold range (0.22–0.79), confirming robustness. Figure~\ref{fig:tri-results}(c) highlights that increasing weights for BLEU and ROUGE-L improves accuracy, emphasizing the role of semantic cues. Overall, \textbf{the Trigger module reliably separates ID/OOD samples with stable performance under varying thresholds and indicator weights}. In subsequent experiments, we fix the threshold at 0.5 with equal weights and observe 100\% ID/OOD separation (see Appendix B.1).

\paragraph{Clustering Analysis of Triggered Buffer.}

\begin{figure*}[t]
    \centering
    \begin{subfigure}[b]{0.25\textwidth}
        \includegraphics[width=\linewidth]{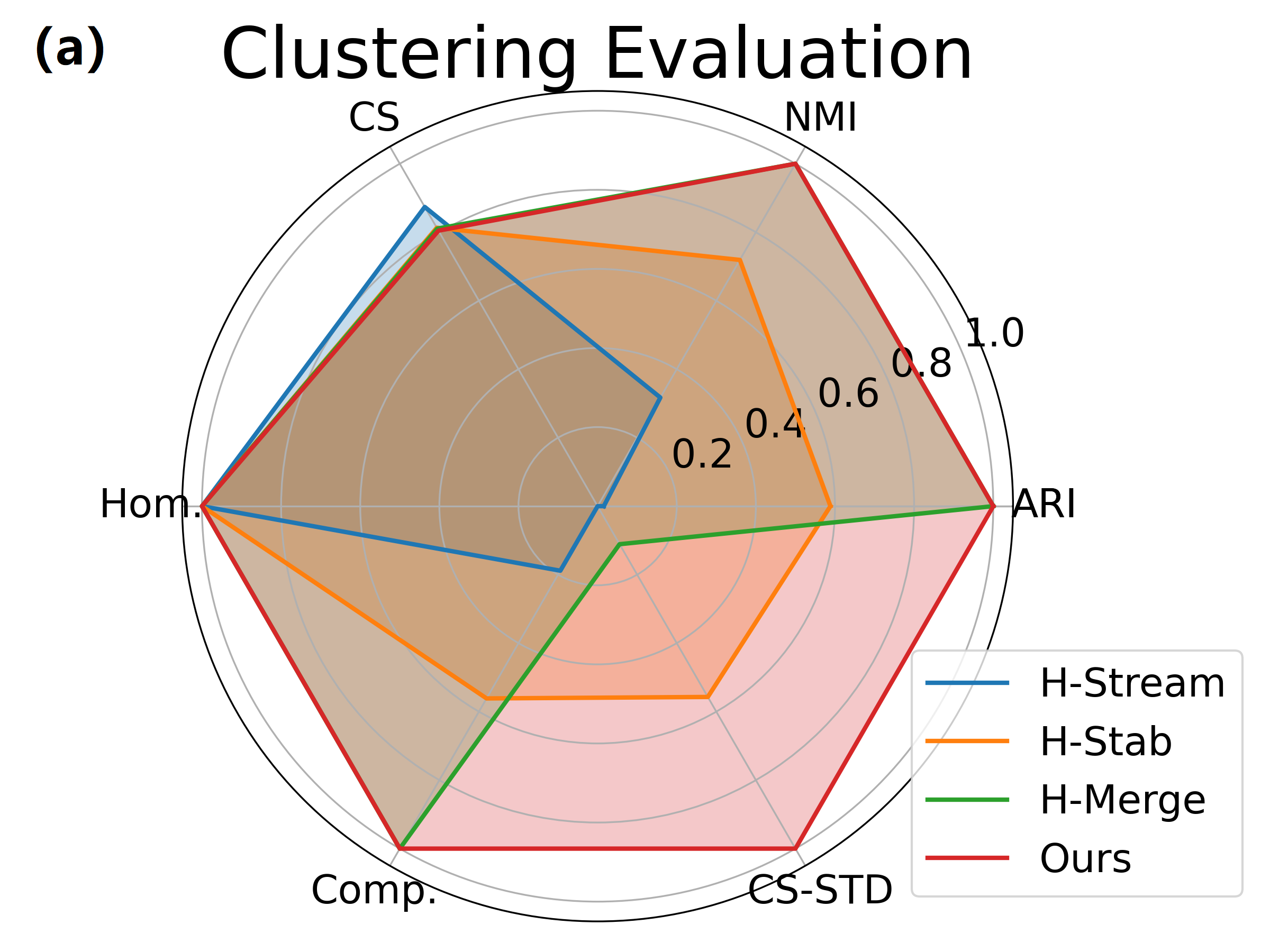}
    \end{subfigure}
    \hfill
    \begin{subfigure}[b]{0.74\textwidth}
        \includegraphics[width=\linewidth]{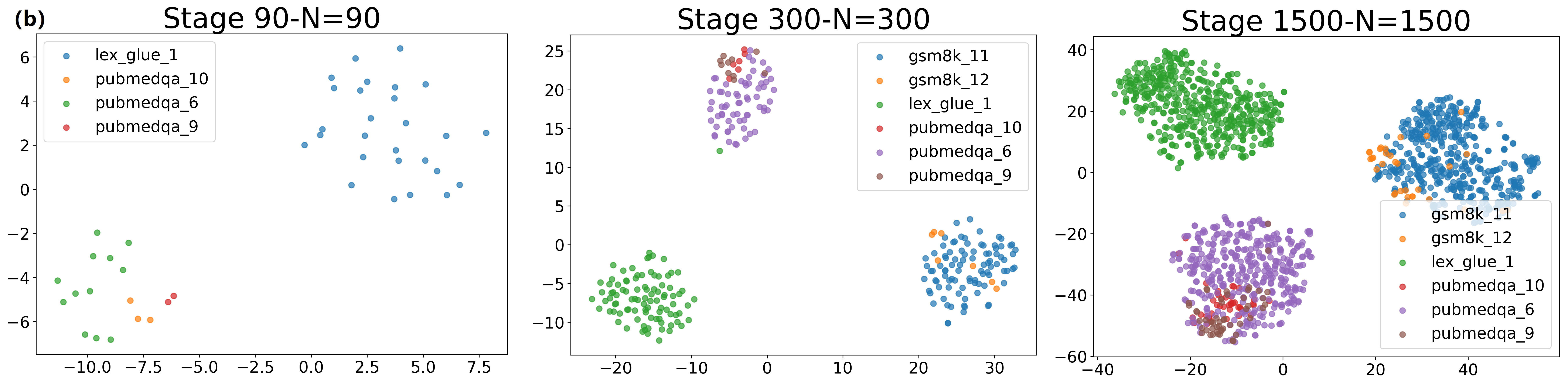}
    \end{subfigure}

    \caption{Results of the Trigger Buffer Module evaluation. (a) Evaluation of Streaming HDBSCAN Variants on Clustering Metrics: Homogeneity, ARI, NMI, Completeness, Central Similarity (CS), and Cluster Sample Std. (CS-STD). (b) Dynamic t-SNE Visualization as Data Volume (N) Increases.}
    \label{fig:tsne}
\end{figure*}
Given the Trigger module's effectiveness, we assume no ID samples entering the Trigger Buffer. Thus, we evaluate its clustering performance using only OOD samples, drawn from the same dataset used in Trigger evaluation.

\begin{figure}[H]
    \centering
    \begin{subfigure}[b]{0.33\textwidth}
        \includegraphics[width=\linewidth]{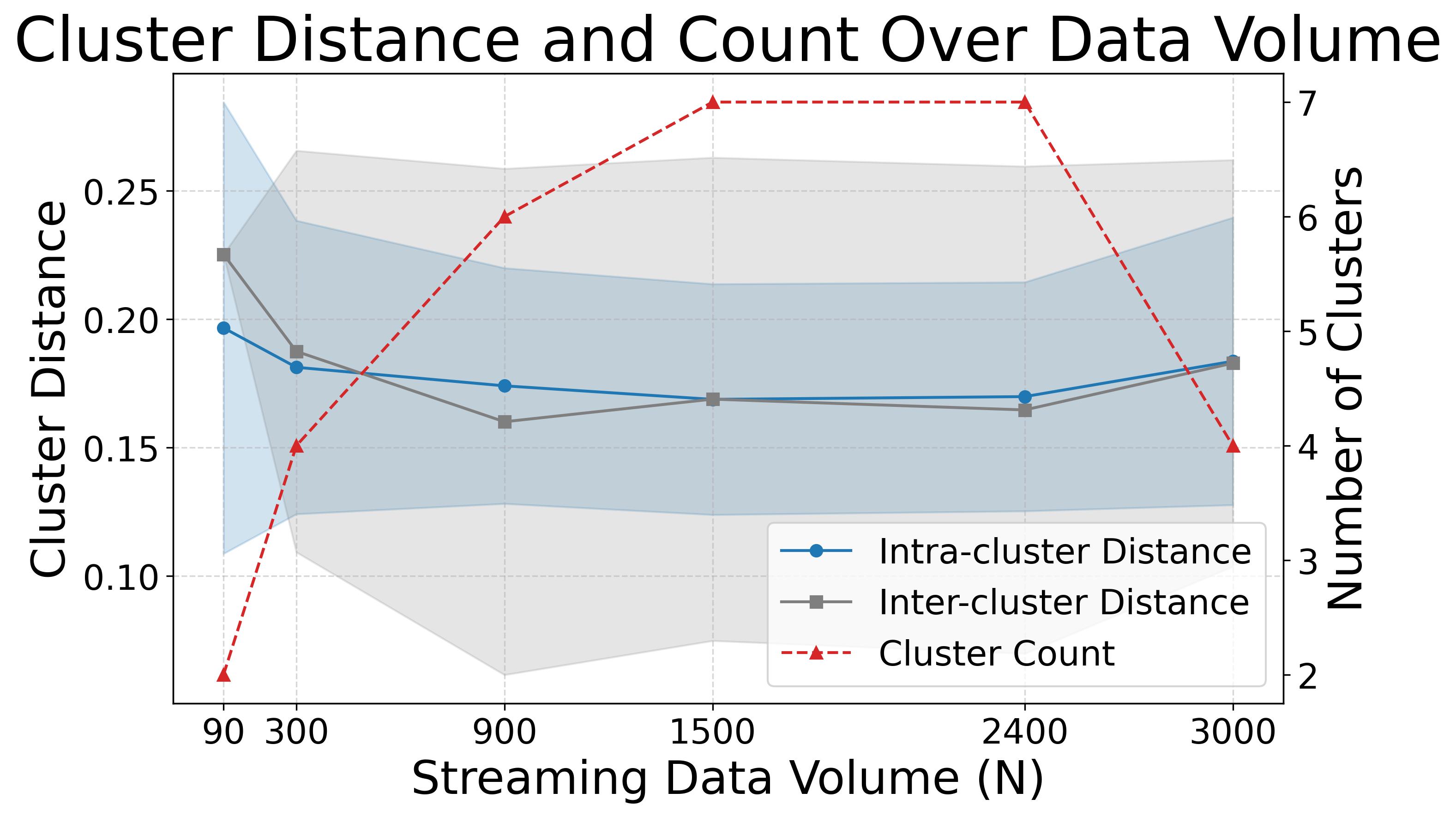}
    \end{subfigure}
    \caption{Cluster distance and cluster count over streaming data volume.}
    \label{fig:clus—d}
\end{figure}

To evaluate the clustering capability of the Trigger Buffer, we conduct three experiments shown in Figure~\ref{fig:tsne} and Figure~\ref{fig:clus—d}. Figure~\ref{fig:tsne}(a) compares six clustering metrics across different strategies. Our SBC algorithm, which combines HDBSCAN, a stability check, and a merge mechanism, achieves the best overall performance. While all methods show good homogeneity due to bucket-based grouping, HDBSCAN alone yields unstable clusters, reflected in low ARI and CS-STD. Adding the stability check improves consistency, and the merge strategy reduces intra-cluster dispersion. SBC algorithm integrates both for consistent and compact clustering. Figure~\ref{fig:tsne}(b) shows clusters becoming more compact as data accumulates. Figure~\ref{fig:clus—d} tracks three metrics: decreasing intra-cluster distance and cluster count indicate redundancy reduction, while stable inter-cluster distance confirms structural robustness. Overall, \textbf{the Trigger Buffer produces compact, stable clusters with convergence behavior in streaming OOD settings.}


\paragraph{Dynamic Adaptation of LoRA Store.}

\begin{table*}[h]
\centering
\begin{threeparttable}
    \begin{tabular}{p{1cm}ccccccccccccc}
        \toprule
        \multicolumn{2}{c}{\multirow{3}{*}{\textbf{Method}}} & \multicolumn{4}{c}{\textbf{PubMedQA(\%)}} & \multicolumn{4}{c}{\textbf{LexGlue(\%)}}  & \multicolumn{4}{c}{\textbf{GSM8K(\%)}}\\
        \cmidrule(r){3-6} \cmidrule(r){7-10} \cmidrule(r){11-14}
        & & \multirow{2}{*}{Acc} & \multicolumn{3}{c}{Macro Avg} & \multirow{2}{*}{Acc} & \multicolumn{3}{c}{Macro Avg} & \multirow{2}{*}{EM} &  \multirow{2}{*}{MAE} & \multirow{2}{*}{MSE} & \multirow{2}{*}{NER} \\
         \cmidrule(r){4-6} \cmidrule(r){8-10}
        & & & P & R & F1 & & P & R & F1 &  &  &  & \\
        \midrule
        \multicolumn{2}{c}{Base Model} & 50.60 & 75.15 & 50.60 & 34.65 & 12.53 & 17.75 & 12.53 & 13.19 & 0 & 175 & 32337 & -\\
        \multicolumn{2}{c}{LoRA$^{*}$} & 67.60 & 75.00 & 67.59 & 65.01 & 65.43 & 73.99 & 85.22 & 65.43 & 20.00 & 481 & $1.8 \times 10^{6}$ & \textbf{100}\\
        \multicolumn{2}{c}{CA-LoRA$^{*}$} & \textbf{84.80} & \textbf{85.71} & \textbf{84.80} & \textbf{84.70} & \textbf{72.21} & \textbf{79.63} & \textbf{72.21} & \textbf{75.50} & 23.38 & 92 & $3.4 \times 10^{8}$ & \textbf{100}\\
        \midrule 
        \multirow{3}{*}{\makecell[{{c}}]{\textbf{$\text{LStore}_{500}$}\\(ours)}}& $\text{Adp}_{1}$ & 43.2 & 48.36 & 43.62 & 34.07 & 34.15 & 27.62 & 34.50 & 34.15 & \textbf{92.32}\cellcolor{gray!20} & \textbf{95}\cellcolor{gray!20} & $8.3 \times 10^{5}$\cellcolor{gray!20} & \textbf{93.99}\cellcolor{gray!20}\\
        & $\text{Adp}_{2}$ & 41.4 & 50.22 & 50.84 & 34.23 & 46.27 & 41.39 & 45.56 & 46.27 & \textbf{81.27}\cellcolor{gray!20} & 473\cellcolor{gray!20} & $3.0 \times 10^{7}$\cellcolor{gray!20} & 63.13\cellcolor{gray!20}\\
        & $\text{Adp}_{3}$ & 58.15 & 50.76 & 51.52 & 50.17 & 36.08 & 38.12 & 35.98 & 36.08 & \textbf{72.13}\cellcolor{gray!20} & 13917\cellcolor{gray!20} & $4.7 \times 10^{9}$\cellcolor{gray!20}  & 57.52\cellcolor{gray!20}\\
        \bottomrule
    \end{tabular}
\end{threeparttable}
\caption{Evaluating the Accuracy of the LoRA Store Module on Diverse Datasets: Highlighting the Role of Atomic Tasks. Symbol \textit{*} denotes the setting where the PubMedQA training set contains 30,000 samples}
\label{biao1}
\end{table*}

\begin{figure}[H]
    \centering
    \begin{subfigure}[b]{0.5\textwidth}
    \includegraphics[width=\linewidth]{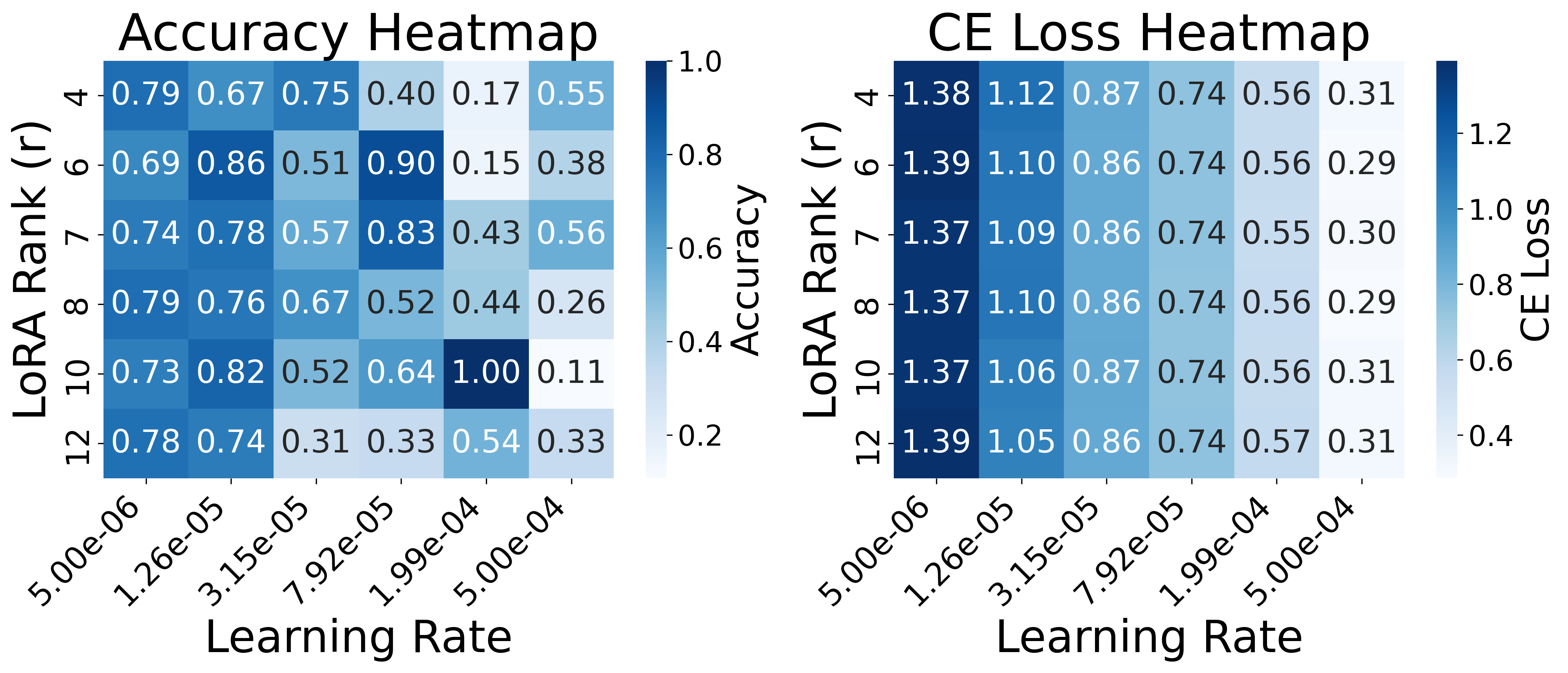}
    \end{subfigure}
    \caption{Heatmap of accuracy and cross-entropy loss for LoRA rank and learning rate: examining the necessity of exploitation optimization.}
    \label{he_lora}
\end{figure}
To assess the necessity of low-rank fine-tuning for atomic tasks, we evaluate the LoRA Store module (Table~\ref{biao1}). Comparisons are made against three baselines: the Base Model (Llama2-7B), standard LoRA (fine-tuned on the full dataset), and CA-LoRA (fine-tuned after static clustering). Results in Table~\ref{biao1} demonstrate the effectiveness of our dynamic adaptation approach in integrating new knowledge. All baseline models used the mixed-task datasets provided. PubMedQA’s and LexGlue tasks enabled strong baseline performance, with CA-LoRA outperforming direct LoRA fine-tuning, highlighting clustering’s benefit. In contrast, GSM8K’s complex arithmetic tasks posed greater challenges, reflected by low exact match (EM) scores despite reasonable numeric extraction (NER). Simple clustering and training yielded limited gains. During LoRA Store training, PubMedQA showed minimal improvement, likely due to task heterogeneity and difficulty in atomic decomposition. However, \textbf{decomposing GSM8K’s level-1 subset (tasks with 1–2 computation steps) significantly boosted reasoning accuracy, validating the effectiveness of atomic tasks}. 

To evaluate the LoRA Store’s effectiveness, heatmaps (Figure~\ref{he_lora}) were created. Accuracy can vary significantly (up to 80\%) and loss decreases depending on the rank and learning rate. Optimal configurations vary by rank, e.g., $r = 6$ at $\text{lr} = 7.92 \times 10^{-5}$, and $r = 10$ at $\text{lr} = 1.99 \times 10^{-4}$. These results highlight \textbf{the necessity of exploitation optimization, as performance varies across rank and learning rate.}

\subsection{End-to-End Evaluation of SAGE}

\begin{table}[h]
\centering
\begin{threeparttable}
    \begin{tabular}{p{1cm}ccccc}
        \toprule
        \multicolumn{2}{c}{\multirow{2}{*}{\textbf{Method}}} &  \multicolumn{4}{c}{\textbf{GSM8K(\%)}} \\
        \cmidrule(r){3-6}
        & & EM & MAE & MSE & NER \\
        \midrule
        \multicolumn{2}{c}{Base Model} & 2.22 & 2258 & $2.0 \times 10^{8}$ & -\\
        \multicolumn{2}{c}{$\text{LoRA}_{499}$} & 7.44 & 2079 & $2.7 \times 10^{8}$ & 99.79 \\
        \multicolumn{2}{c}{$\text{CA-LoRA}_{484}$} & 10.64 & 1096 & $8.1 \times 10^{7}$ & 99.80 \\
        \midrule 
        \multirow{3}{*}{\makecell[{{c}}]{\textbf{$\text{LStore}_{499}$}\\(ours)}}& $\text{Adp}_{1}$ & 92.32 & 95 & $8.3 \times 10^{5}$ & 93.99\\
        & $\text{Adp}_{2}$ & 81.27 & 473 & $3.0 \times 10^{7}$ & 63.13\\
        & $\text{Adp}_{3}$ & 72.13 & 13917 & $4.7 \times 10^{9}$ & 57.52\\
        \midrule 
        \multirow{3}{*}{\makecell[{{c}}]{\textbf{$\text{SAGE}_{315}$}\\(ours)}}& $\text{Adp}_{1}$ & \textbf{84.95} & 28185 & $1.5 \times 10^{10}$ & 78.58\\
        & $\text{Adp}_{2}$ & \textbf{99.80} & 0.05 & 1.36 & 99.40 \\
        & $\text{Adp}_{3}$ & \textbf{99.80$\uparrow$} & \textbf{0.01$\downarrow$} & \textbf{0.16$\downarrow$} & \textbf{100$\uparrow$} \\
        \rowcolor{gray!20}
        \multicolumn{3}{l}{\textbf{Random Seed of EM}} &  \multicolumn{3}{r}{\textbf{$97.16_{\pm 4.65}$}}\\
        \bottomrule
    \end{tabular}
\end{threeparttable}
\caption{Evaluating the Accuracy of the LoRA Buffer Module on Diverse Datasets: Highlighting the Role of Atomic Tasks. The $\uparrow$ indicates the maximum value, and the $\downarrow$ indicates the minimum value.}
\label{biao2}
\end{table}

We evaluated SAGE by integrating all three modules, focusing on the GSM8K Level-1 dataset due to LoRA Store's superior performance on atomic tasks. Three baselines were used: Base Model, LoRA, and CA-LoRA. Final results, including LoRA Store fine-tuning, are shown in Table~\ref{biao2}. Results show that SAGE with dynamic clustering outperforms the unclusterd LoRA Store. We further performed the Wilcoxon signed-rank test to assess statistical significance. \textbf{The result ($W = 0.0, p = 0.0039$) confirms that the improvement of SAGE over the baseline is statistically significant ($p < 0.01$).} Clustering refines the dataset, reduces noise, and improves convergence with fewer samples. During evaluation, the Top-3 adapters were retained and the effectiveness of this approach was verified. While all reached 100\% training accuracy, only $\text{Adapter}_{3}$ $\text{Adp}_{3}$ generalized well, achieving the best test EM, MSE, MAE, and digit extraction rate. To assess stability, we shuffled the data with different seeds. As shown in Appendix B.2, \textbf{SAGE's performance remained consistent, indicating strong robustness to data order}. Also, the shaded rows in Table~\ref{biao2} report the mean and variance of EM scores over 9 adapters from the random seed experiments.

\subsection{Ablation Study}
\begin{table}[h]
\centering
\setlength{\tabcolsep}{1mm}
\begin{threeparttable}
    \begin{tabular}{cccccc}
        \toprule
        \multicolumn{2}{c}{\multirow{2}{*}{\textbf{Method}}} &  \multicolumn{4}{c}{\textbf{GSM8K(\%)}} \\
        \cmidrule(r){3-6}
        & & EM & MAE & MSE & NER \\
        \midrule 
        \multirow{3}{*}{\makecell[{{c}}]{\textbf{$\text{LStore}_{499}$}\\(random)}}& $\text{Adp}_{1}$ & 6.44 & 11609 & $3.6 \times 10^{10}$ & 99.17\\
        & $\text{Adp}_{2}$ & 8.89 & 11594 & $3.6 \times 10^{10}$ & 100\\
        & $\text{Adp}_{3}$ & 6.72 & 14995 & $3.9 \times 10^{10}$ & 99.17 \\
        \midrule
        \multirow{3}{*}{\makecell[{{c}}]{\textbf{$\text{LStore}_{499}$}\\(level-1)}}& $\text{Adp}_{1}$ & 92.32 & 95 & $8.3 \times 10^{5}$ & 93.99\\
        & $\text{Adp}_{2}$ & 81.27 & 473 & $3.0 \times 10^{7}$ & 63.13\\
        & $\text{Adp}_{3}$ & 72.13 & 13917 & $4.7 \times 10^{9}$ & 57.52\\
        \midrule
        \multirow{3}{*}{\makecell[{{c}}]{\textbf{$\text{HLStore}_{484}$}\\(HDBSCAN)}}& $\text{Adp}_{1}$ & 89.77 & 309 & $1.9 \times 10^{7}$ & 98.97\\
        & $\text{Adp}_{2}$ & 78.70 & 21812 & $9.6 \times 10^{9}$ & 89.26\\
        & $\text{Adp}_{3}$ & 96.69 & 429 & $8.9 \times 10^{7}$ & 100 \\
        \midrule
        \multirow{3}{*}{\makecell[{{c}}]{\textbf{$\text{KLStore}_{329}$}\\(K-means)}}& $\text{Adp}_{1}$ & 97.55 & 14 & 31717 & 99.09\\
        & $\text{Adp}_{2}$ & 74.47 & 377 & $6.5 \times 10^{5}$ & 100\\
        & $\text{Adp}_{3}$ & 92.40 & 646 & $1.3 \times 10^{8}$ & 100 \\
        \midrule 
        \multirow{3}{*}{\makecell[{{c}}]{\textbf{$\text{SAGE}_{315}$}\\(\textbf{ours})}}& $\text{Adp}_{1}$ & 84.95 & 28185 & $1.5 \times 10^{10}$ & 78.58\\
        & $\text{Adp}_{2}$ & \textbf{99.80} & \textbf{0.05} & \textbf{1.36} & \textbf{99.40}\\
        & $\text{Adp}_{3}$ & \textbf{99.80$\uparrow$} & \textbf{0.01$\downarrow$} & \textbf{0.16$\downarrow$} & \textbf{100$\uparrow$} \\
        \bottomrule
    \end{tabular}
\end{threeparttable}
\caption{Ablation Study of SAGE on the GSM8K Dataset.}
\label{biao4}
\end{table}
We omit ablation studies on the Trigger and LoRA Store modules, as their functionalities have been validated in prior experiments: the Trigger module reliably distinguishes ID and OOD data to control dynamic inference, while the LoRA Store handles dynamic hyperparameter tuning, outperforming static baselines. Our ablation focuses on the Trigger Buffer and atomic task datasets. As shown in Table~\ref{biao4}, we compare following baseline settings: (1) LoRA Store without task decomposition or clustering, (2) with task decomposition but no Trigger Buffer, and (3) with static clustering (HLStore, KLStore). \textbf{SAGE outperforms all variants in EM, MAE, and NER, confirming the effectiveness of atomic task splitting and dynamic clustering via the Trigger Buffer.}

\section{Related Work}
\paragraph{Self-Adaptive Mechanisms in LLMs}
Self-adaptive mechanisms are considered essential for memory-augmented LLMs. Early work identified key-value-like storage behavior in Transformer FFNs~\cite{geva2021transformer}, inspiring enhanced memory routing~\cite{wumemorizing, berges2024memory, li2025cmt}, long-context inputs~\cite{wang2023augmenting}, and retrieval-augmented pathways~\cite{borgeaud2022improving} to improve knowledge persistence. Memory$\textsuperscript{3}$\cite{lewis2020retrieval, liu2024much} spurred interest in retrieval-augmented generation for knowledge updating. More recently, MemoryLLM~\cite{2024MEMORYLLM} shifted focus to self-adaptive mechanisms, enabling LLMs to autonomously revise internal representations—a move from passive storage to active self-adaptation~\cite{zhong2022training}.

\paragraph{LoRA-Based Adaptation and Reusability}
Concurrently, LoRA and its variants have advanced parameter-efficient fine-tuning~\cite{he2021towards}. Drawing inspiration from sparse mixture-of-experts (MoE)\cite{arnob2025exploring}. Increasing focus has been placed on LoRA’s modularity and reusability~\cite{ostapenko2024towards, xu2024meteora, citation-0, valipourdylora, liao2025dynamic}, with methods like Switch-LoRA~\cite{kong2024lora} supporting dynamic adapter selection and plug-and-play use. Optimizing LoRA training efficiency, including hyperparameter tuning and update latency, remains an active research direction.

\section{Conclusion and Future Work}
We introduced the SAGE framework, which segments reasoning tasks into atomic tasks and dynamically fine-tunes LoRA parameters through failure-triggered data detection and clustering. This approach enables dynamic adaptation at test-time, improving comprehension on new knowledge. However, the current design of the Trigger module may be limited by dataset alignment, which could affect the user experience. Future work could explore replacing the Trigger module with neural networks to enhance flexibility and scalability. Additionally, while the atomic task approach works well for simpler tasks, it requires further development to address more complex reasoning tasks.

\bibliography{aaai2026}

\begin{thebibliography}{31}
\providecommand{\natexlab}[1]{#1}

\bibitem[{Berges et~al.(2024)Berges, O{\u{g}}uz, Haziza, Yih, Zettlemoyer, and Gosh}]{berges2024memory}
Berges, V.-P.; O{\u{g}}uz, B.; Haziza, D.; Yih, W.-t.; Zettlemoyer, L.; and Gosh, G. 2024.
\newblock Memory Layers at Scale.
\newblock \emph{arXiv e-prints}, arXiv--2412.

\bibitem[{Borgeaud et~al.(2022)Borgeaud, Mensch, Hoffmann, Cai, Rutherford, Millican, Van Den~Driessche, Lespiau, Damoc, Clark et~al.}]{borgeaud2022improving}
Borgeaud, S.; Mensch, A.; Hoffmann, J.; Cai, T.; Rutherford, E.; Millican, K.; Van Den~Driessche, G.~B.; Lespiau, J.-B.; Damoc, B.; Clark, A.; et~al. 2022.
\newblock Improving language models by retrieving from trillions of tokens.
\newblock In \emph{International conference on machine learning}, 2206--2240. PMLR.

\bibitem[{Campello, Moulavi, and Sander(2013)}]{campello2013density}
Campello, R.~J.; Moulavi, D.; and Sander, J. 2013.
\newblock Density-based clustering based on hierarchical density estimates.
\newblock In \emph{Pacific-Asia conference on knowledge discovery and data mining}, 160--172. Springer.

\bibitem[{Dziri et~al.(2023)Dziri, Lu, Sclar, Li, Jiang, Lin, Welleck, West, Bhagavatula, Le~Bras et~al.}]{dziri2023faith}
Dziri, N.; Lu, X.; Sclar, M.; Li, X.~L.; Jiang, L.; Lin, B.~Y.; Welleck, S.; West, P.; Bhagavatula, C.; Le~Bras, R.; et~al. 2023.
\newblock Faith and fate: Limits of transformers on compositionality.
\newblock \emph{Advances in Neural Information Processing Systems}, 36: 70293--70332.

\bibitem[{Geva et~al.(2021)Geva, Schuster, Berant, and Levy}]{geva2021transformer}
Geva, M.; Schuster, R.; Berant, J.; and Levy, O. 2021.
\newblock Transformer Feed-Forward Layers Are Key-Value Memories.
\newblock In \emph{Proceedings of the 2021 Conference on Empirical Methods in Natural Language Processing}. Association for Computational Linguistics.

\bibitem[{He et~al.(2021)He, Zhou, Ma, Berg-Kirkpatrick, and Neubig}]{he2021towards}
He, J.; Zhou, C.; Ma, X.; Berg-Kirkpatrick, T.; and Neubig, G. 2021.
\newblock Towards a Unified View of Parameter-Efficient Transfer Learning.
\newblock \emph{arXiv e-prints}, arXiv--2110.

\bibitem[{Hu et~al.(2022)Hu, Shen, Wallis, Allen-Zhu, Li, Wang, Wang, Chen et~al.}]{hu2022lora}
Hu, E.~J.; Shen, Y.; Wallis, P.; Allen-Zhu, Z.; Li, Y.; Wang, S.; Wang, L.; Chen, W.; et~al. 2022.
\newblock Lora: Low-rank adaptation of large language models.
\newblock \emph{ICLR}, 1(2): 3.

\bibitem[{Jin et~al.(2025)Jin, Zhang, Luo, and Wang}]{jin2025reasoning}
Jin, H.; Zhang, P.; Luo, M.; and Wang, H. 2025.
\newblock Reasoning Can Hurt the Inductive Abilities of Large Language Models.
\newblock \emph{arXiv preprint arXiv:2505.24225}.

\bibitem[{Kil et~al.(2024)Kil, Mai, Lee, Chowdhury, Wang, Cheng, Wang, Liu, and Chao}]{kil2024mllm}
Kil, J.; Mai, Z.; Lee, J.; Chowdhury, A.; Wang, Z.; Cheng, K.; Wang, L.; Liu, Y.; and Chao, W.-L.~H. 2024.
\newblock Mllm-compbench: A comparative reasoning benchmark for multimodal llms.
\newblock \emph{Advances in Neural Information Processing Systems}, 37: 28798--28827.

\bibitem[{Kong et~al.(2024)Kong, Li, Fang, Feng, He, Dong, Wang, Li, Kong, and Liu}]{kong2024lora}
Kong, R.; Li, Q.; Fang, X.; Feng, Q.; He, Q.; Dong, Y.; Wang, W.; Li, Y.; Kong, L.; and Liu, Y. 2024.
\newblock LoRA-Switch: Boosting the Efficiency of Dynamic LLM Adapters via System-Algorithm Co-design.
\newblock \emph{CoRR}.

\bibitem[{Lewis et~al.(2020)Lewis, Perez, Piktus, Petroni, Karpukhin, Goyal, K{\"u}ttler, Lewis, Yih, Rockt{\"a}schel et~al.}]{lewis2020retrieval}
Lewis, P.; Perez, E.; Piktus, A.; Petroni, F.; Karpukhin, V.; Goyal, N.; K{\"u}ttler, H.; Lewis, M.; Yih, W.-t.; Rockt{\"a}schel, T.; et~al. 2020.
\newblock Retrieval-augmented generation for knowledge-intensive nlp tasks.
\newblock \emph{Advances in neural information processing systems}, 33: 9459--9474.

\bibitem[{Li et~al.(2025)Li, Sun, Hu, Hu, and Zhang}]{li2025cmt}
Li, D.; Sun, Z.; Hu, X.; Hu, B.; and Zhang, M. 2025.
\newblock Cmt: A memory compression method for continual knowledge learning of large language models.
\newblock In \emph{Proceedings of the AAAI Conference on Artificial Intelligence}, volume~39, 24413--24421.

\bibitem[{Li, Han, and Shihao(2024)}]{citation-0}
Li, Y.; Han, S.; and Shihao, J. 2024.
\newblock VB-LoRA: Extreme Parameter Efficient Fine-Tuning with Vector Banks.
\newblock In \emph{The Thirty-eighth Annual Conference on Neural Information Processing Systems}.

\bibitem[{Liao et~al.(2025)Liao, Wang, Zhou, Hu, Zheng, and Gao}]{liao2025dynamic}
Liao, X.; Wang, C.; Zhou, S.; Hu, J.; Zheng, H.; and Gao, J. 2025.
\newblock Dynamic Adaptation of LoRA Fine-Tuning for Efficient and Task-Specific Optimization of Large Language Models.
\newblock \emph{arXiv preprint arXiv:2501.14859}.

\bibitem[{Lin(2004)}]{lin2004rouge}
Lin, C.-Y. 2004.
\newblock Rouge: A package for automatic evaluation of summaries.
\newblock In \emph{Text summarization branches out}, 74--81.

\bibitem[{Lin et~al.(2017)Lin, Feng, dos Santos, Yu, Xiang, Zhou, and Bengio}]{lin2017structured}
Lin, Z.; Feng, M.; dos Santos, C.~N.; Yu, M.; Xiang, B.; Zhou, B.; and Bengio, Y. 2017.
\newblock A STRUCTURED SELF-ATTENTIVE SENTENCE EMBEDDING.
\newblock In \emph{International Conference on Learning Representations}.

\bibitem[{Liu, Lin, and Liu(2024)}]{liu2024much}
Liu, J.; Lin, J.; and Liu, Y. 2024.
\newblock How much can rag help the reasoning of llm?
\newblock \emph{arXiv preprint arXiv:2410.02338}.

\bibitem[{Liu et~al.(2016)Liu, Wen, Yu, and Yang}]{liu2016large}
Liu, W.; Wen, Y.; Yu, Z.; and Yang, M. 2016.
\newblock Large-margin softmax loss for convolutional neural networks.
\newblock In \emph{Proceedings of the 33rd International Conference on International Conference on Machine Learning-Volume 48}, 507--516.

\bibitem[{Ostapenko et~al.(2024)Ostapenko, Su, Ponti, Charlin, Le~Roux, Caccia, and Sordoni}]{ostapenko2024towards}
Ostapenko, O.; Su, Z.; Ponti, E.~M.; Charlin, L.; Le~Roux, N.; Caccia, L.; and Sordoni, A. 2024.
\newblock Towards modular LLMs by building and reusing a library of LoRAs.
\newblock In \emph{Proceedings of the 41st International Conference on Machine Learning}, 38885--38904.

\bibitem[{Ouyang et~al.(2022)Ouyang, Wu, Jiang, Almeida, Wainwright, Mishkin, Zhang, Agarwal, Slama, Ray et~al.}]{ouyang2022training}
Ouyang, L.; Wu, J.; Jiang, X.; Almeida, D.; Wainwright, C.; Mishkin, P.; Zhang, C.; Agarwal, S.; Slama, K.; Ray, A.; et~al. 2022.
\newblock Training language models to follow instructions with human feedback.
\newblock \emph{Advances in neural information processing systems}, 35: 27730--27744.

\bibitem[{Papineni et~al.(2002)Papineni, Roukos, Ward, and Zhu}]{papineni2002bleu}
Papineni, K.; Roukos, S.; Ward, T.; and Zhu, W.-J. 2002.
\newblock Bleu: a method for automatic evaluation of machine translation.
\newblock In \emph{Proceedings of the 40th annual meeting of the Association for Computational Linguistics}, 311--318.

\bibitem[{Shazeer et~al.(2017)Shazeer, Mirhoseini, Maziarz, Davis, Le, Hinton, and Dean}]{shazeer2017outrageously}
Shazeer, N.; Mirhoseini, A.; Maziarz, K.; Davis, A.; Le, Q.; Hinton, G.; and Dean, J. 2017.
\newblock Outrageously large neural networks: The sparsely-gated mixture-of-experts layer.
\newblock \emph{arXiv preprint arXiv:1701.06538}.

\bibitem[{Tie et~al.(2025)Tie, Zhao, Song, Wei, Zhou, Dai, Yin, Yang, Yan, Su et~al.}]{tie2025survey}
Tie, G.; Zhao, Z.; Song, D.; Wei, F.; Zhou, R.; Dai, Y.; Yin, W.; Yang, Z.; Yan, J.; Su, Y.; et~al. 2025.
\newblock A survey on post-training of large language models.
\newblock \emph{arXiv e-prints}, arXiv--2503.

\bibitem[{Valipour et~al.(2022)Valipour, Rezagholizadeh, Kobyzev, and Ghodsi}]{valipourdylora}
Valipour, M.; Rezagholizadeh, M.; Kobyzev, I.; and Ghodsi, A. 2022.
\newblock DyLoRA: Parameter Efficient Tuning of Pre-trained Models using Dynamic Search-Free Low Rank Adaptation.
\newblock \emph{NeurIPS}.

\bibitem[{Wang et~al.(2023)Wang, Dong, Cheng, Liu, Yan, Gao, and Wei}]{wang2023augmenting}
Wang, W.; Dong, L.; Cheng, H.; Liu, X.; Yan, X.; Gao, J.; and Wei, F. 2023.
\newblock Augmenting language models with long-term memory.
\newblock \emph{Advances in Neural Information Processing Systems}, 36: 74530--74543.

\bibitem[{Wang et~al.(2024)Wang, Gao, Chen, Jiang, Li, Yang, Yin, Li, Li, and Yin}]{2024MEMORYLLM}
Wang, Y.; Gao, Y.; Chen, X.; Jiang, H.; Li, S.; Yang, J.; Yin, Q.; Li, Z.; Li, X.; and Yin, B. 2024.
\newblock MEMORYLLM: Towards Self-Updatable Large Language Models.
\newblock \emph{ICML}.

\bibitem[{Wu et~al.(2022)Wu, Rabe, Hutchins, and Szegedy}]{wumemorizing}
Wu, Y.; Rabe, M.~N.; Hutchins, D.; and Szegedy, C. 2022.
\newblock Memorizing Transformers.
\newblock In \emph{International Conference on Learning Representations}.

\bibitem[{Xu, Lai, and Huang(2024)}]{xu2024meteora}
Xu, J.; Lai, J.; and Huang, Y. 2024.
\newblock MeteoRA: Multiple-tasks Embedded LoRA for Large Language Models.
\newblock \emph{arXiv e-prints}, arXiv--2405.

\bibitem[{Yang et~al.(2024)Yang, Lin, Wang, Wu, Li, Tang, Wei, Wang, Tang, Song et~al.}]{yang2024memory3}
Yang, H.; Lin, Z.; Wang, W.; Wu, H.; Li, Z.; Tang, B.; Wei, W.; Wang, J.; Tang, Z.; Song, S.; et~al. 2024.
\newblock Memory3: Language modeling with explicit memory.
\newblock \emph{arXiv preprint arXiv:2407.01178}.

\bibitem[{Yeasar et~al.(2025)Yeasar, Su, Kim, Ostapenko, Precup, Caccia, and Alessandro}]{arnob2025exploring}
Yeasar, A.~S.; Su, Z.; Kim, M.; Ostapenko, O.; Precup, D.; Caccia, L.; and Alessandro, S. 2025.
\newblock Exploring Sparse Adapters for Scalable Merging of Parameter Efficient Experts.
\newblock In \emph{ICLR 2025 Workshop on Modularity for Collaborative, Decentralized, and Continual Deep Learning}.

\bibitem[{Zhong, Lei, and Chen(2022)}]{zhong2022training}
Zhong, Z.; Lei, T.; and Chen, D. 2022.
\newblock Training Language Models with Memory Augmentation.
\newblock In \emph{2022 Conference on Empirical Methods in Natural Language Processing, EMNLP 2022}.

\end{thebibliography}
\section{Appendix}
\subsection{A. Implementation Details}
\subsubsection{A.1 Details for Trigger Buffer Module of SBC Algorithm}
The Trigger Buffer module of Streaming Buffer Clustering (SBC) algorithm details are as follows:

\begin{algorithm}[ht]
\caption{Streaming Buffer Clustering (SBC)}
\label{alg:structure_cluster}
\begin{algorithmic}[1]
    \Require Incoming anomaly sample $x$, data buffer $\mathcal{B}$, current clusters $\mathcal{C}$, structure tag $s$
    \Ensure Cluster assignment or data buffer update

    \State $s \gets \text{InferStructure}(x)$
    \State $e \gets \text{GetEmbedding}(x)$ 
    \State $k \gets \text{ExtractKeywords}(x)$ 

    \If{IsClusterStable(s) == True}
        \State $(\mathcal{C}^*, \gamma) \gets \text{FindBestCluster}(s, e, k)$  \\  \textcolor{gray}{\# $\gamma$: weighted score combining embedding similarity $e$ and keyword overlap $k$}
        \If{$\gamma \geq \tau$}   \Comment{Exceeding threshold $\tau$}
            \State Assign $x$ to cluster $\mathcal{C}^*$
            \State \Return cluster assignment
        \EndIf
    \Else
        \State \text{AddToBuffer}$(x, e, k, \mathcal{B}_s)$
    \EndIf

    \If{$|\mathcal{B}_s| \geq T$}   \Comment{Exceeding clustering threshold $T$}
        \State $e \gets \text{ExtractEmbeddings}(\mathcal{B}_s)$
        \State $h \gets \text{HDBSCAN}(e)$ \\ \Comment{Density clustering without preset count}
        \State $(\text{ARI}, \text{Sim}) \gets \text{EvalStability}(h, h^{(t-1)}, e, e^{(t-1)})$
        \If{$\text{ARI} \geq \theta_{\text{ari}} \land \text{Sim} \geq 
        \theta_{\text{sim}}$}
            \State \text{CreateNewClusters}($h, \mathcal{B}_s$)
            \State $\mathcal{B}_s\gets \emptyset$ \Comment{Clear buffer}
            \State $\text{StableFlag}_s\gets \text{True}$
            \If{$|\mathcal{C}_s| \geq 3$}
                \State \text{MergeSimilarClusters}($\mathcal{C}_s$) \\
                \Comment{Merge clusters with centroid similarity}
            \EndIf
        \Else
            \State \Return \Comment{Clustering unstable, wait more}
        \EndIf
    \EndIf
    
    \State \Return $x$ as ``Unassigned'' \Comment{No matching cluster}
\end{algorithmic}
\end{algorithm}

\subsubsection{A.2 Details for LoRA Store Module of CLO Algorithm}

The LoRA Store module of Cluster-Aware LoRA Optimization (CLO) algorithm, a clustering-based method that automatically searches for the optimal LoRA configuration $\theta^*$, saving the corresponding adapter for subsequent inference or combination, details are as follows:

\vspace*{-\topskip}
\begin{algorithm}[t]
\caption{Cluster-Aware LoRA Optimization (CLO)}
\label{alg:lora_optimizer}
\begin{algorithmic}[1]
    \Require Incoming stable cluster $\mathcal{C}$, base model$\mathcal{M}$, tokenizer $\mathcal{T}$, parameter space $\mathcal{P}$, top-$k$ configs $k$
    \Ensure Optimized LoRA config $\theta^{*}$ and LoRA adapter$\mathcal{A}$

    \State $\Theta_0 \gets \text{InitialConfigs}(\mathcal{P}, n)$ \Comment{Initialize $n$ seed configs}
    \State $\mathcal{R}_0 \gets []$ \Comment{Initial training results}

    \ForAll{$\theta \in \Theta_0$}
        \State $r \gets \text{TrainLoRA}(\theta, \mathcal{C}, \mathcal{M}, \mathcal{T})$
        \State $\mathcal{R}_0 \gets \mathcal{R}_0 \cup \{r\}$
    \EndFor

    \State $\mathcal{R}_0 \gets \text{SortByScore}(\mathcal{R}_0)$ \\ 
    \textcolor{gray}{\# maximize accuracy, break ties by minimizing loss}
    \State $\Theta_{\text{top}} \gets \text{Top-KConfigs}(\mathcal{R}_0, k)$
    
    \State $\mathcal{R}_1 \gets []$ \Comment{Refined training results}
    \ForAll{$\theta_{\text{top}} \in \Theta_{\text{top}}$}
        \State $\Theta_{\text{opt}} \gets \text{LocalParamSearch}(\theta_{\text{top}})$ \\
        \textcolor{gray}{\#local search over rank and learning rate near $\theta_{\text{top}}$}
        \ForAll{$\theta' \in \Theta_{\text{opt}}$}
            \State $r' \gets \text{TrainLoRA}(\theta', \mathcal{C}, \mathcal{M}, \mathcal{T})$
            \State $\mathcal{R}_1 \gets \mathcal{R}_1 \cup \{r'\}$
        \EndFor
    \EndFor

    \State $\mathcal{R}_{\text{all}} \gets \mathcal{R}_0 \cup \mathcal{R}_1$
    \State $\mathcal{R}_{\text{all}} \gets \text{SortByScore}(\mathcal{R}_{\text{all}})$
    \State $\mathcal{A}_{\text{LoRA}} \gets \text{Top-KResults}(\mathcal{R}_{\text{all}}, 3)$
    \State $\text{SaveBuffer}(\mathcal{A}_{\text{LoRA}})$
    
    \State \Return $\mathcal{A}_{\text{LoRA}}$ and $\theta^*$ 
\end{algorithmic}
\end{algorithm}

\subsection{B. Experimental Details}

\subsection{B.1 Details for Trigger Module}
In further experiments, we fixed the threshold at 0.5 and set equal weights for all four indicators. We then evaluated the Trigger module independently under these test-time parameters. As shown in Figure~ref{fig:tri-sj}, the module achieved 100\% ID/OOD separation under this configuration, confirming that the chosen parameters fall within a favorable operational range. Together with the sensitivity analysis in Figure~\ref{fig:tri-results}, these findings suggest that the Trigger module possesses not only high accuracy but also strong robustness and generalizability, making it well-suited to serve as a reliable activation mechanism for subsequent adaptive fine-tuning in LLM-based systems.

\begin{figure}[H]
    \centering
    \begin{subfigure}[b]{0.23\textwidth}
        \includegraphics[width=\linewidth]{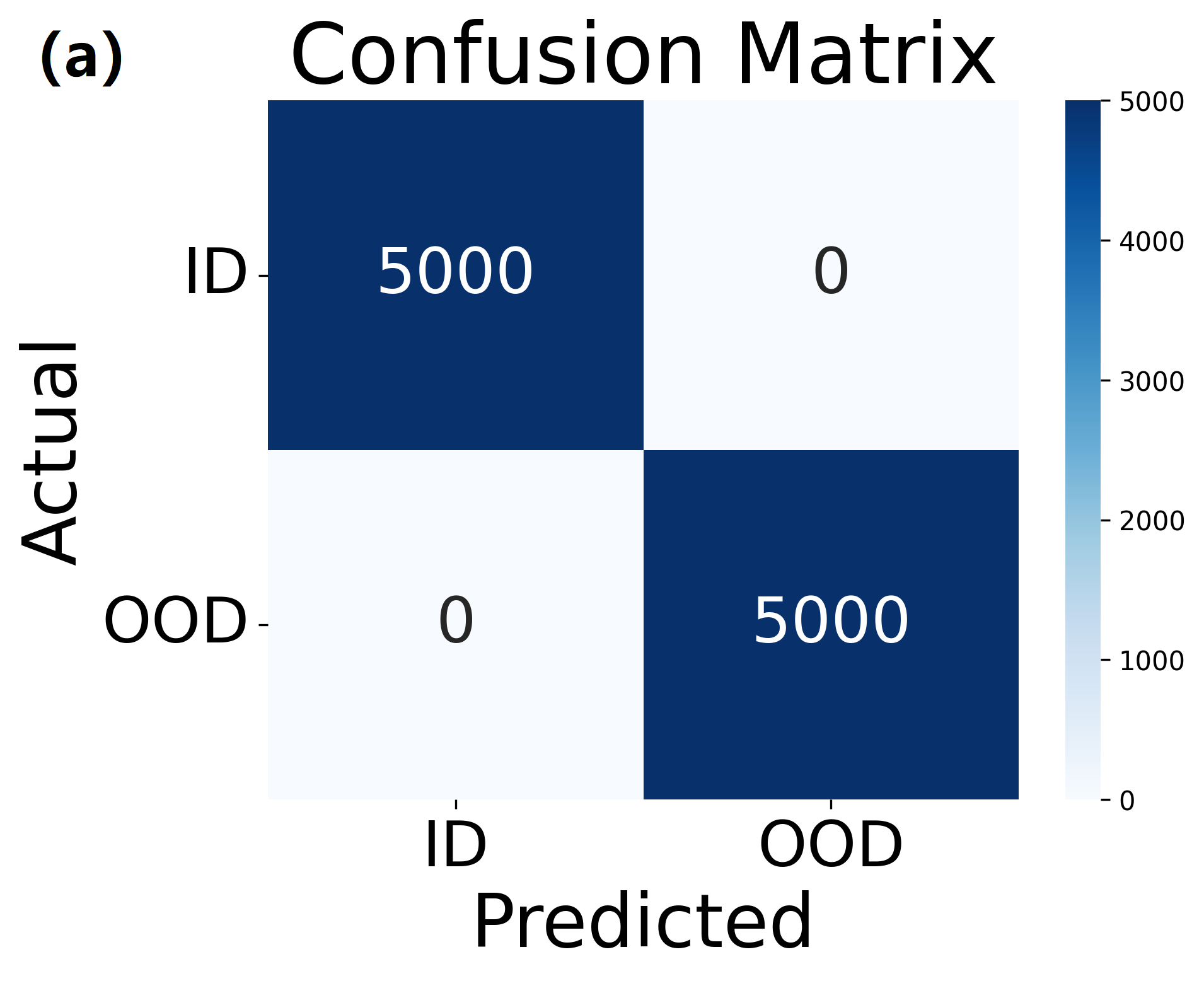}
    \end{subfigure}
    \hfill
    \begin{subfigure}[b]{0.23\textwidth}
        \includegraphics[width=\linewidth]{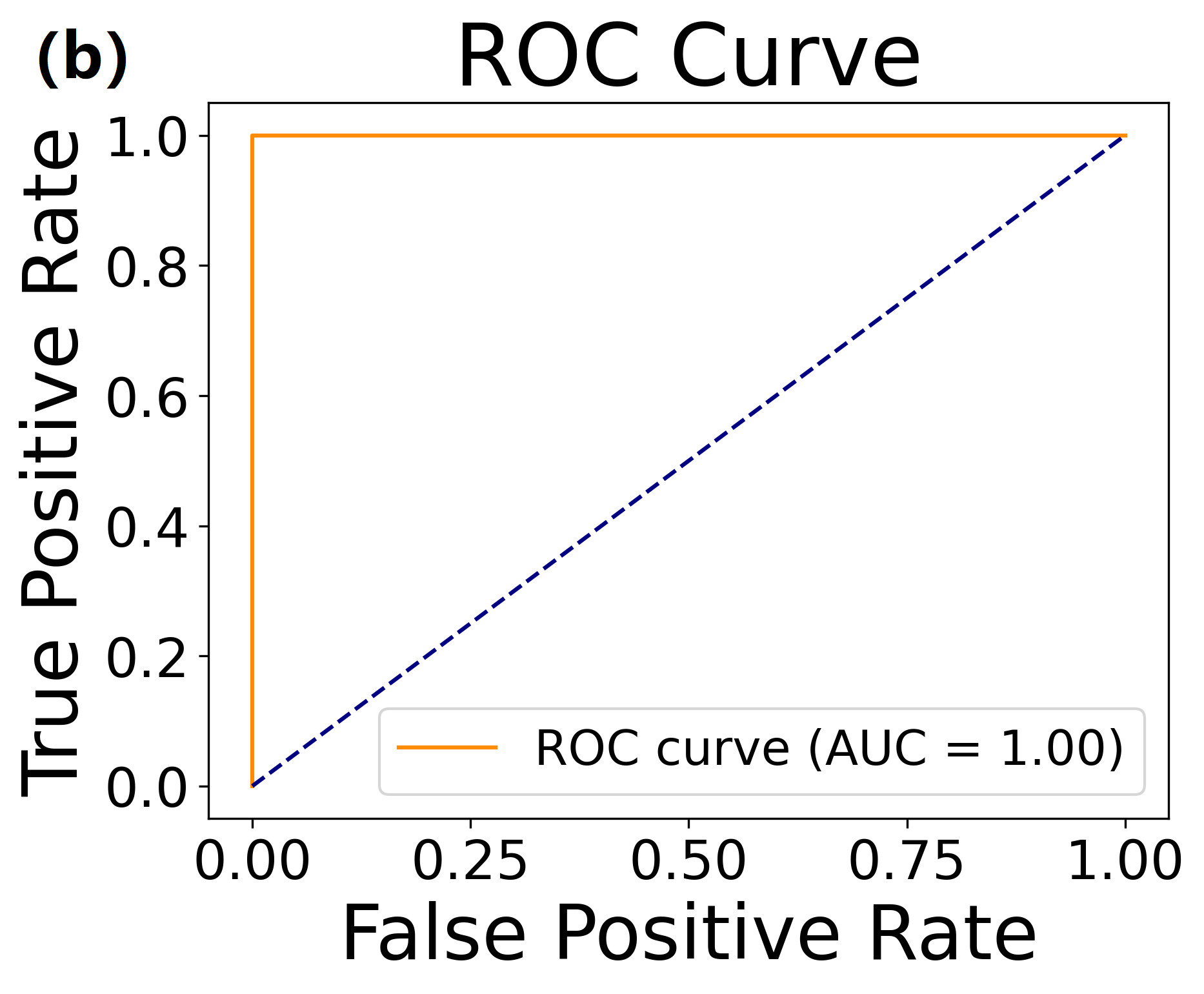}
    \end{subfigure}

    \caption{Results of the Trigger Module evaluation under formal test parameters. (a) Confusion matrix for ID and OOD samples. (b) ROC curve illustrating detection performance.}
    \label{fig:tri-sj}
\end{figure}

\subsection{B.2 Details for SAGE}

To further assess the stability of SAGE, we tested the model by shuffling the data using different random seeds. The final results are presented in Table 3. The evaluation indicates that shuffling the data order has no significant impact on the performance of SAGE, demonstrating its high robustness to changes in data order.
\begin{table*}[t]
\centering
\label{biao3}
\begin{threeparttable}
    \begin{tabular}{ccccccccccccc}
        \toprule
        \multirow{2}{*}{Seed} & \multicolumn{4}{c}{123} & \multicolumn{4}{c}{42} & \multicolumn{4}{c}{7}\\
        \cmidrule(r){2-5}\cmidrule(r){6-9}\cmidrule(r){10-13}
        &EM & MAE & MSE & NER  &EM & MAE & MSE & NER  &EM & MAE & MSE & NER\\
        \midrule 
        $\text{Adp}_{1}$ & 84.95 &29185 & $1.5 \times 10^{10}$ & 78.58 & 98.28 & 11 & 39277& 81.36 & 94.92 & 547 & $9.6 \times 10^{7}$ & 94.59\\
        $\text{Adp}_{2}$ & 99.80 &0.05 & 1.36 & 99.40 & 95.63 & 1561 & $5.1 \times 10^{8}$ & 96.39 & 99.80 & 0.01 & 0.05 &100\\
        $\text{Adp}_{3}$ & 99.80 & 0.01 & 0.16 & 100 & 99.60 & 0.05 & 1.36 & 100 & 98.99 & 4.56 & 5705 &99.00\\
        \bottomrule
    \end{tabular}
\end{threeparttable}
\caption{Impact of Random Seed Variation on SAGE}
\end{table*}

\section{C. Related Work}
\subsection{Self-Adapting Mechanisms in LLMs}
Since the emergence of LLMs, a series of studies have investigated memory mechanisms inspired by cognitive systems, particularly viewing LLMs as brain-like architectures. Early work observed that the feed-forward networks (FFNs) in Transformer models exhibit key-value-like storage behavior~\cite{geva2021transformer}, prompting efforts to enhance memory functionality by routing information through these components~\cite{wumemorizing, berges2024memory, li2025cmt}. This line of research also motivated the construction of long-context~\cite{wang2023augmenting} inputs and retrieval-enhanced pathways~\cite{borgeaud2022improving} to improve internal knowledge persistence. A significant milestone was marked by the introduction of Memory$\textsuperscript{3}$~\cite{yang2024memory3}, which explicitly decomposed memory into three components: model parameters, explicit memory modules, and retrieved text. The study demonstrated that externalized memory not only facilitates faster knowledge acquisition but also serves as a viable mechanism for persistent knowledge representation in LLMs. This perspective helped establish the notion that memory accelerates learning and catalyzed a surge of interest in RAG~\cite{lewis2020retrieval, liu2024much} for knowledge updating and memory simulation. More recently, the development of MemoryLLM~\cite{2024MEMORYLLM} has shifted attention toward mechanisms for self-updating. Rather than solely focusing on constructing long-term or short-term memory banks, this work highlights the importance of enabling LLMs to selectively and autonomously revise their internal representations—marking~\cite{zhong2022training} a conceptual shift from passive storage to active self-modification.

\subsection{LoRA-Based Adaptation and Reusability}
In parallel with memory research, substantial progress has been made in advancing LoRA and its variants for parameter-efficient fine-tuning~\cite{he2021towards}. Inspired in part by the success of sparse mixture-of-experts (MoE)~\cite{shazeer2017outrageously} architectures, recent studies have proposed sparsity-aware LoRA designs to support efficient and selective adaptation in data-limited settings~\cite{arnob2025exploring}. Beyond sparsity, increasing attention has been given to the modularity and reusability~\cite{ostapenko2024towards, xu2024meteora, citation-0, valipourdylora, liao2025dynamic} of LoRA modules. Approaches such as Switch-LoRA~\cite{kong2024lora} explore dynamic adapter selection and plug-and-play capabilities, enabling rapid composition and reuse across diverse tasks. Meanwhile, the training efficiency of LoRA—particularly in terms of hyperparameter optimization and update latency—has become an active area of research.

\end{document}